\documentclass[times,twocolumn,final]{elsarticle}

\usepackage{prletters}
\usepackage{framed,multirow}
\usepackage{float}

\usepackage{amssymb}
\usepackage{latexsym}

\usepackage{url}
\usepackage{xcolor}
\usepackage{amsmath}
\usepackage{tabularx}
\usepackage{comment}
\usepackage{caption}
\usepackage{subcaption}
\definecolor{newcolor}{rgb}{.8,.349,.1}

\usepackage[inline]{enumitem}

\usepackage{titlesec}

\journal{Pattern Recognition Letters}

\titlespacing*{\section}{0pt}{7pt}{7pt}
\titlespacing*{\subsection}{0pt}{7pt}{7pt}
\titlespacing*{\subsubsection}{0pt}{7pt}{7pt}

\begin{document}

\thispagestyle{empty}

\clearpage

\ifpreprint
  \setcounter{page}{1}
\else
  \setcounter{page}{1}
\fi

\begin{frontmatter}

\title{Towards Practical Implementations of Person Re-Identification from Full Video Frames}

\author[1]{Felix O. \snm{Sumari} H.}

\author[1]{Luigy \snm{Machaca}}
\author[1]{Jose \snm{Huaman}}
\author[1]{Esteban W. G. \snm{Clua}} 
\author[2]{Joris \snm{Gu\'erin}\corref{cor1}}
\cortext[cor1]{Corresponding author: 
  Tel.: +55 (84) 99616-3888;}
\ead{jorisguerin.research@gmail.com}

\address[1]{Universidade Federal Fluminense, Instituto de Computa\c{c}\~ao, Niter\'oi-RJ, 24320-150, BRAZIL}
\address[2]{Universidade Federal do Rio Grande do Norte, Departamento de Engenharia de Computa\c{c}\~ao e Automa\c{c}\~ao, Natal-RN, 59078-970, BRAZIL}

\received{1 May 2013}
\finalform{10 May 2013}
\accepted{13 May 2013}
\availableonline{15 May 2013}
\communicated{S. Sarkar}

\begin{abstract}

With the major adoption of automation for cities security, person re-identification (Re-ID) has been extensively studied recently. In this paper, we argue that the current way of studying person re-identification, i.e. by trying to re-identify a person within already detected and pre-cropped images of people, is not sufficient to implement practical security applications, where the inputs to the system are the full frames of the video streams. To support this claim, we introduce the Full Frame Person Re-ID setting (FF-PRID) and define specific metrics to evaluate FF-PRID implementations. To improve robustness, we also formalize the hybrid human-machine collaboration framework, which is inherent to any Re-ID security applications. To demonstrate the importance of considering the FF-PRID setting, we build an experiment showing that combining a good people detection network with a good Re-ID model does not necessarily produce good results for the final application. This underlines a failure of the current formulation in assessing the quality of a Re-ID model and justifies the use of different metrics. We hope that this work will motivate the research community to consider the full problem in order to develop algorithms that are better suited to real-world scenarios.
\end{abstract}


\end{frontmatter}


\section{Introduction}
\label{sec:intro}

\vspace{-5pt}
In recent years, many security cameras were deployed in public places such as streets, malls or airports. Today, most of these video streams are monitored in real-time by security agents, which is expensive and rather inefficient as the amount of videos to analyze is tremendous. In contrast, automated video analysis~\cite{hampapur2003smart} can process large amounts of videos simultaneously but is more prone to errors for complex tasks such as 
person re-identification~\cite{ye2020deep}. In addition, even for automated video analysis systems, the final decision often rests with a human security agent, who triggers the appropriate actions. Hence, in practice it seems good to adopt hybrid approaches, where artificial intelligence models can screen the whole network in real time and select only relevant sequences for the monitoring agents.

In computer vision, the person Re-Identification (Re-ID) problem aims at searching a given person (query) in a network of non-overlapping cameras and raising an alert when this person appears in one of the video streams. It seeks to reproduce and enhance the human ability to recognize people in different scenarios, e.g. wearing different clothes, 
in a different pose, etc. 

\begin{figure}[ht]
\centering
\includegraphics[width=0.35\textwidth]{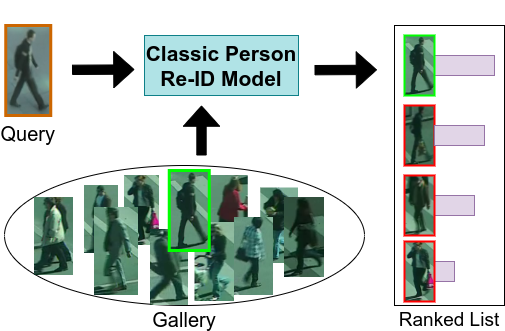}
\caption{Illustration of the Classic Person Re-ID (C-PRID) setting.}
\label{fig:classic_reid}
\end{figure}

The current formulation to address Re-ID (Fig.~\ref{fig:classic_reid}) is based on large databases of images representing human beings in a real-world environment~\cite{market, duke, cuhk, viper, prid}. 
These images are usually extracted using pedestrian detection models~\cite{pedestrian_detection} and filtered manually to meet certain standards: each image should contain the entire body of exactly one person, centered and occupying most of the image. From these datasets, a given image is selected as the query and the others constitute the search gallery. Then, the objective is to look for the query person within the gallery \cite{ye2020deep}. 
This approach is illustrated in Fig.~\ref{fig:classic_reid}. Sometimes, individual images are replaced by sequences of successive cropped images and the problem is called video-based Re-ID~\cite{PRL_videoBased, PRL_videoBased2}. From now on, the Re-ID setting considering pre-cropped images of persons as input is referred to as Classic Person Re-Identification (C-PRID). Recent successful methods to address C-PRID are mostly based on deep learning~\cite{Ejaz2015, Wang2018, Zheng2019, PRL_multiLevelAttention, PRL_ensemble_reid}.



\begin{figure}[ht]
\centering
\includegraphics[width=0.45\textwidth ]{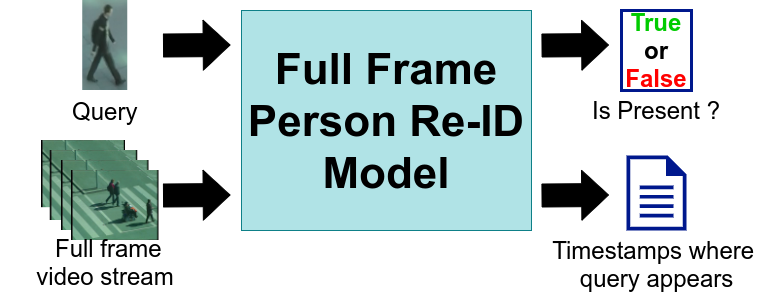}
\caption{Illustration of the Full Frame Person Re-ID (FF-PRID) setting.}
\label{fig:realWorld_reid}
\end{figure}

This formulation is a useful building block to implement security application of Re-ID, i.e. to identify a person sought by the authorities in a network of security cameras. However it is not sufficient, as for such a practical implementation, the entire image of the video frames must be used as input, instead of carefully selected pre-cropped images of persons. From now on, this application-oriented Re-ID setting is referred to as Full Frame Person Re-ID (FF-PRID). 
In short, in the FF-PRID setting, a successful model must analyze full frames to determine if the query is present in the stream, and if it is, when and where it appeared. The FF-PRID setting is illustrated in Fig.~\ref{fig:realWorld_reid}.

\vspace{5pt}

One can argue that the C-PRID problem can be easily derived from the FF-PRID setting by applying a pedestrian detection~(PD) model~\cite{pedestrian_detection} on the raw video stream, which is often done in practice. Indeed, some object detection models have demonstrated strong results for detecting human beings over the last few years~\cite{ren2015faster, yolov3, pedestrian_detection2}. However, we argue that not considering the problem as a whole presents several issues:
\vspace{-5pt}
\begin{itemize}[leftmargin=*]
\setlength\itemsep{0.001em}
    \item The bounding boxes extracted by PD models may differ from the images in the reference datasets used for C-PRID training and evaluation, which have been filtered manually to only select clean images. This domain shift between the galleries used for training and the data encountered at inference time can decrease the quality of the model at run time, and thus induces a strong bias for model evaluation. 

    \item Even if both a good pedestrian detection model and a good Re-ID model are used, their small prediction errors might add up to produce poor overall results for the final application.
    
    \item Not considering FF-PRID as an independent problem might dissuade the community from trying different approaches for the full application. Indeed, the vast availability of C-PRID datasets might take researchers away from trying other promising approaches such as end-to-end methods or video based methods, which have been shown to work for other computer vision problems~\cite{end2end_1, video_based_1}.
    
    \item When developing a practical application, it is crucial to evaluate the quality of the entire pipeline before deploying it in production. To the best of our knowledge, frameworks and metrics to evaluate FF-PRID are missing in the literature.
\end{itemize}{}
\vspace{-5pt}

The main contributions of this paper is to introduces a new setting of the person Re-ID problem, called FF-PRID, which is better suited to implement and evaluate real-world security applications. By formalizing the natural collaboration occurring between an automated Re-ID system and the monitoring agents, a hybrid framework to address the FF-PRID problem more robustly is proposed, as well as two complementary metrics to assess the quality of any FF-PRID pipeline. Then, experiments are conducted to demonstrate the importance of considering the FF-PRID problem in its entirety. The most natural pipeline for FF-PRID is implemented within the proposed framework, it consists in using a pedestrian detection model and a C-PRID model sequentially, with both models performing well on standard datasets for their respective tasks. This FF-PRID pipeline is then tested on a modified version of the PRID-2011 dataset~\cite{prid}, using the metrics introduced in this paper. Our experimental results demonstrate that this combination struggles to produce good results for the FF-PRID problem, despite the apparent success of its two independent components. This shows the importance of considering the person Re-ID problem in its Full Frame setting, using adapted metrics. This research does not claim to have solved the FF-PRID setting but rather to demonstrate that the current way of approaching Re-ID is not suited for practical scenarios. By proposing an alternative framework and evaluation method, we hope that this work will motivate the community to consider the FF-PRID setting, in order to develop algorithms that are better adapted for real-world scenarios.
\section{Human-Machine Hybrid Framework for FF-PRID}
\label{sec:reformulation}

The classic Re-ID formulation consists in comparing a query image with all the images of a search gallery to output a set of similarity scores representing the Re-ID predictions. Conversely, this paper considers the Full Frame Re-ID setting, which is better suited to implement and evaluate security applications. In practice, security application involving Re-ID are inherently human-machine collaboration tasks as the predictions made by a FF-PRID model are used by security agents to decide the actions to be taken. In this section, this interaction is formalized within a hybrid framework to increase robustness of FF-PRID pipelines. We also propose two new evaluation metrics to assess the quality of a FF-PRID model.

\subsection{Framework}
\label{sec:framework}
A FF-PRID system takes as input both a raw video and a query image. Studying this setting is paramount to design reliable Re-ID applications. In the ideal scenario, a FF-PRID model should make predictions for every single frame, and raise an alert as soon as the query is encountered. A monitoring agent can then analyze this alert and trigger appropriate actions. However, in practice FF-PRID for security applications is usually a critical task, which is complex and highly prone to errors. To address this issue, we propose to formalize the interaction occurring naturally between the Re-ID system and the monitoring agent. This way, by introducing user defined parameters to relax the ideal scenario, the amount of human work can be tuned to obtain the required robustness. The ideal case discussed above is a special case of the generic framework presented here.  

\begin{figure*}
\centering

\includegraphics[width=\textwidth ]{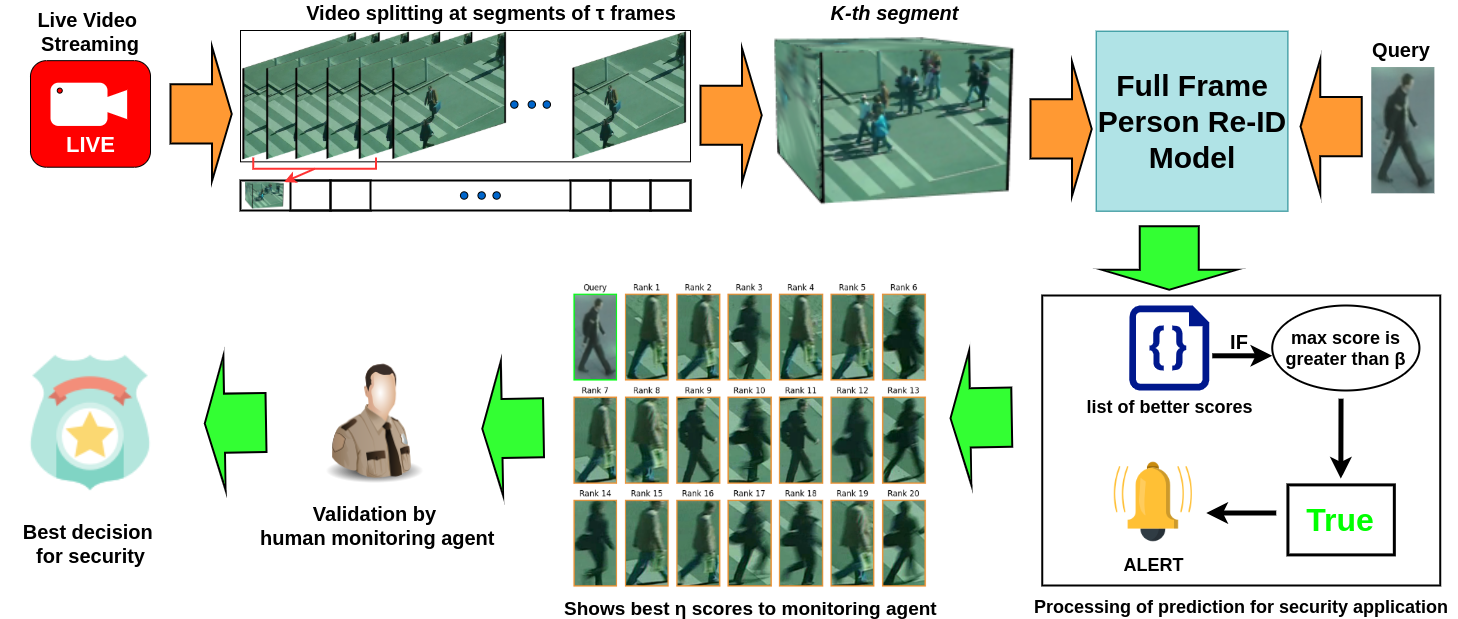}
\caption{Hybrid Human-Machine FF-PRID Framework.}

\label{fig:ffprid-framework}
\end{figure*}

The proposed framework goes as follows: First, the live video stream is cut into short video segments of $\tau$ frames. Then, a FF-PRID model processes each video segment, and outputs a set of cropped images of persons present in the video, as well as their corresponding similarity scores with respect to the query image. If the highest similarity score returned by the model is higher than a given threshold $\beta$, the $\eta$ members of the gallery with highest similarity scores are shown to the monitoring agent, who decides if the predictions are correct and triggers actions when necessary. An overview of this hybrid framework is shown in Figure~\ref{fig:ffprid-framework}. We underline  that the only assumption made regarding the FF-PRID model is about the shape of its inputs and outputs. A practical example of FF-PRID model is studied in our experiments (Section~\ref{sec:experiments}).  

The threshold for raising an alert $\beta$, the number of images shown to the agent $\eta$ and the length of the video segments $\tau$ are user defined parameters that influence the final results. For any real-world implementation of Re-ID for city security, it is always required to tune $\beta$, independently of our proposed framework. The two additional parameters $\tau$ and $\eta$ were introduced because we acknowledged that in practice, the ideal scenario described above ($\tau=1$ and $\eta=1$) generates very poor results.

\subsection{Evaluation metric}

For a perfect FF-PRID model, the operator validation is required in all the cases where the query is present in the sequence of $\tau$~frames and not in any other case. Hence, there are two ways for a model to fail: by missing the query when it is present in the video segment or by calling the operator when the query is not present. Thus, to evaluate the quality of a model, we define two important indicators that we call \emph{Finding Rate} (FR) and \emph{True Validation Rate} (TVR). They respectively represent the number of sequences in which the query was found when it appeared and the number of times that the query was present when the operator was solicited. 

To define FR and TVR formally, other variables must be introduced first. For a given $\{$query, video$\}$ pair, we define:
\vspace{-5pt}
\begin{itemize}[leftmargin=*]
\setlength\itemsep{0.001em}
    \item A True Call (TC), when the query is present in the video, the highest similarity score is greater than the threshold $\beta$ and the query is in the top $\eta$ best candidates. It corresponds to a successful case of re-identification by the system.
    \item a True Missed Call (TMC), when the query is present in the video, the highest similarity score is greater than $\beta$ and the query is not in the top $\eta$ best candidates. It is the case where the query is present, the system is asking for confirmation but does not provide the correct images to the operator and the query is missed anyways.
    \item a False Silence (FS), when the query is present in the video, but the highest similarity score is smaller than $\beta$. It is the case where the query is missed but the operator is not disturbed.
    \item a False Call (FC), when the query is not in the video but the highest similarity score is greater than $\beta$. It corresponds to the case where the operator is disturbed for nothing.
    \item a True Silence (TS), when the query is not in the video and the highest similarity score is smaller than $\beta$. It is the case where the query is not present and nothing happens.
\end{itemize}{}
Then, the proposed metrics can be defined as follows:
\begin{equation}
    FR = \frac{TC}{TC + TMC + FS},
\end{equation}{}
\begin{equation}
    TVR = \frac{TC}{TC + TMC + FC}.
\end{equation}{}
FR and TVR are comprised between 0 and 1. When FR~=~1, it means that whenever the query was present in the video, it was successfully identified by the hybrid system (model + operator). Likewise, TVR~=~1 means that the operator was never called unnecessarily, i.e. the query was present in the proposed images whenever the model asked for verification. In contrast, FR~$<$~1 means that in some sequences the query was missed, and TVR~$<$~1 means that validation was required in cases where the query was not among the suggestions. 


\section{Proposed experiments}
\label{sec:experiments}

To demonstrate the importance of considering the FF-PRID setting when implementing Re-ID in real-world security applications, we implement a very natural FF-PRID pipeline and evaluate it under the proposed metrics. 

\subsection{Dataset used for our experiments}

For these experiments, a modified version of the PRID-2011 dataset~\cite{prid} is used, considering full frame videos instead of the pre-cropped, manually filtered images of the original dataset.

The original PRID-2011 dataset is composed of images extracted from multiple person trajectories recorded from two different static surveillance cameras, named A and B. Images from these cameras contain a view point change and a stark difference in illumination and background. Since images are extracted from trajectories, several successive poses per person are available in each camera view, with some people appearing in both views. 
After filtering out manually some heavily occluded persons, corrupted images induced by tracking and annotation errors, the official PRID-2011 dataset contains 385 persons in camera view A and 749 in camera view B. The persons with the first 200 labels appear in both views. 

PRID-2011 was created to test classic Re-ID approaches, as well as video-based Re-ID~\cite{PRL_videoBased}. To conduct our experiments, we obtained the raw videos and annotations used to create the PRID-2011 dataset\footnote{We thank the authors of the PRID-2011 paper for their cooperation.}. From now on, the two raw full frame videos are called view~A (1:01:53 hours) and view~B (1:06:39 hours). Both views were cut into sub-videos of 2 minutes, to serve as input to the FF-PRID model (Section~\ref{sec:expe_model}). This way, view~A contains 30 videos and view~B contains 33. 


For each 2 minute video sample, we generate a ground truth file\footnote{Our code (video processing script and baseline model implementation) is openly available at: \url{https://github.com/fsumari/FF-PRID-2020}.}, which contains containing for each person in the video: its identifier $Id$; the first frame where it appears $fr$; the number of frames in which it appears $s$ and the initial bounding box coordinates $(ulx, uly, brx, bry)$.

\subsection{FF-PRID model used for our experiments}
\label{sec:expe_model}

\begin{figure*}
\centering
\includegraphics[width=0.95\textwidth ]{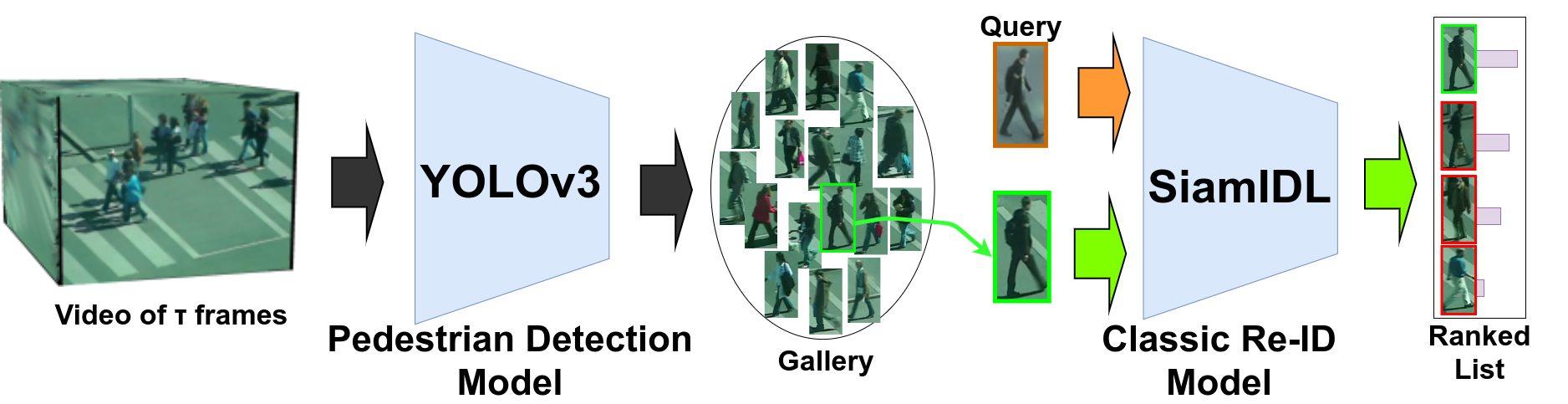}

\vspace{3pt}
\caption{Standard pipeline for FF-PRID used in our experiments.}

\label{fig:pipeline-rw-reid-modified}
\end{figure*}

The FF-PRID model used in our experiments is presented in Fig.~\ref{fig:pipeline-rw-reid-modified}. The input video is fed to an Object Detection (OD) model to detect pedestrians and generate clippings for the search gallery. Next, the query is searched in the gallery by means of a C-PRID model, returning a list of images, ordered from most to least similar to the query. Currently, this pipeline is to the most common approach to implementing a real-world application of Re-ID. Both the PD model and the C-PRID model were implemented using TensorFlow~1.14.0 and executed on a NVIDIA P5000 GPU. The implementation of these models is detailed in the following subsections.

\subsection{Object Detection}


In this paper, we use the You Only Look Once (YOLO-v3)~\cite{yolov3} approach for PD. In short, YOLO methods belong to the family of regression/classification based approaches, mapping directly from image pixels to bounding box coordinates and class probabilities to reduce significantly the time complexity. A detailed explanation of YOLO is out of the scope of this paper, and for a complete overview of the recent literature about Object Detection (OD), we refer the reader to the two following surveys~\cite{OD_survey, OD_survey2}.

In practice, we use the Darknet-53 convolutional architecture and pretrained weights proposed in tensorflow. 
This network uses 53 convolutional layers with $3x3$ kernels in the beginning and $1x1$ in the end. The model used was trained on the VOC dataset~\cite{voc}, containing 80 classes. Darknet-53 operates at a level close to state-of-the-art object detectors, but is faster because it uses less floating-point operations. The YOLO-v3 model was prepared with a threshold of 0.5 for both Intersection over union (IOU) and the loss function. During our evaluation (Section~\ref{sec:results_OD}), the score threshold to keep a bounding box, as well as the IOU threshold were both set to 0.5 as well. To generate the search galleries, we only use the output corresponding to the person class from the object detector.



\subsection{Classic Person Re-ID}




In our experiments, we perform C-PRID using a method called SiamIDL~\cite{Ejaz2015}. It uses a Siamese neural network architecture composed of two layers of tied convolution with max pooling, cross-input neighborhood differences, patch summary features, across-patch features, higher-order relationships, and finally a softmax function to yield the final estimate of whether the input images are of the same person or not.

For implementation, we used the source code provided by the authors and trained the network using the training set of the CUHK-03 dataset~\cite{cuhk}, containing 7239 images. We use the same parameters as in the original paper: batch\_size=50, max\_steps=210 000, and learning\_rate=0.01. To evaluate the model, the Cumulative Matching Characteristics (CMC) are computed on both the validation folder of CUHK-03 (938 images) and the original PRID-2011 dataset.

\section{Results}
\label{sec:results}

To demonstrate the importance of evaluating Re-ID models in the full frame setting, and thus corroborate the usefulness of the proposed metrics, the evaluation conducted here is two-fold. First, both the PD and the C-PRID models are evaluated independently on standalone datasets. Then, the full FF-PRID pipeline is evaluated using our proposed metrics.

\subsection{Evaluation of the Pedestrian Detection model}
\label{sec:results_OD}

The PRID-2011 dataset was initially created to evaluate C-PRID models. 
Hence, occluded persons, persons with less than five confidence frames, as well as distorted images caused by tracking and annotation errors were removed from the list of Bounding Boxes (BB) (see Figure~\ref{fig:imagePRID}). 
To achieve a correct evaluation of YOLO-v3 on the PRID-2011 videos, it is necessary to manually add the BBs of these people who were ignored during dataset creation. To do this, the \textit{LabelIMG} tool was used and a total of 37 772 BBs were added to the labels of video B. The results obtained for pedestrian detection with YOLO-v3 on PRID-2011 are presented in Table~\ref{table:resultsEvaluationOD}. These results correspond to the model used to generate the search gallery for the C-PRID model (Fig.~\ref{fig:pipeline-rw-reid-modified}).\par

\begin{table}[ht]
\caption{Evaluation of YOLO-v3 pedestrian detection model on PRID-2011 (video B). For the Original BBs (OBB), only the ground truth labels from the original dataset are used. For the OBB + Manually added BBs (MBB), the BBs added manually are also considered.}
\begin{tabular}{c|c|c|c|c}
\multicolumn{1}{c|}{}   & \textbf{Precision} & \textbf{Recall} & \textbf{F1-score} & \textbf{mAP} \\ \hline
\textbf{OBB}     & 0.462      & 0.866      & 0.603      & 45.53\%      \\ \hline
\textbf{OBB + MBB} & 0.761    &  0.824     & 0.791       & 69.50\%     \\ 
\end{tabular}
\label{table:resultsEvaluationOD}
\end{table}

\begin{figure}[ht]
     \centering
     \includegraphics[width=0.4\textwidth]{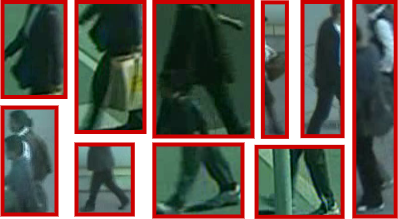}
     \caption{Illustration of the different mistakes for cropping}
     \label{fig:mistakesCropps}
\end{figure}

When displaying the BBs predicted by YOLO-v3 on PRID-2011 video~B, the results look almost perfect. This way, the difference in the results between OBB and OBB+MBB can be interpreted as the number of entire human bodies which where manually filtered to obtain the original dataset (e.g. partially overlapping persons). The remaining errors for the OBB+MBB case mostly correspond to incomplete body parts, such as legs, arms or torso (Fig.~\ref{fig:mistakesCropps}), which were not included in our ground truth BBs.
An object detector, such as YOLO-v3, is trained to find characteristics of the object of interest and thus generates BBs for the cases mentioned above. This constitutes an important discrepancy between the domain on which the C-PRID model was trained and the images generated by the PD model. Such domain shift in the inputs of the C-PRID model can be a major source of errors for the complete FF-PRID pipeline.

\subsection{Evaluation of the classic Re-ID model}
\label{sec:results_REID}

\begin{figure}[ht]
     \centering
     \includegraphics[width=0.4\textwidth]{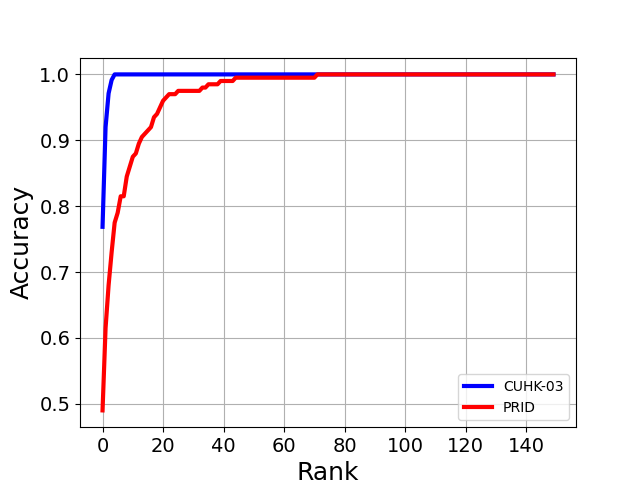}
     \caption{CMC curve on CUHK-03 validation set and on view A of PRID-2011, using a SiamIDL model trained on CUHK-03 training set.}
     \label{fig:rank}
\end{figure}

To evaluate the SiamIDL model used in our experiments, we compute the CMC curves for both the validation set of CUHK-03 and PRID-2011. These results can be seen on Fig.~\ref{fig:rank}. The evaluation on CUHK-03 is used to validate our training by comparing the results obtained with the ones reported in the original paper. The blue curve in Fig.~\ref{fig:rank} is very similar to the experimental results obtained in~\cite{Ejaz2015}. The red curve on Fig.~\ref{fig:rank} shows the results of a test performed on the first 200 Ids from view A of PRID-2011. We can see that the results obtained were good, with more than 48\% on Rank 1 and more than 95\% on Rank 20. No additional training was conducted on the PRID-2011 dataset and only the weights trained on CUHK-03 were used in this validation. This last experiment evaluates the practical scenario of deploying Re-ID in new environments (e.g. new city, new shopping center), where it would be impractical to create a new custom training dataset for every new implementation.

\begin{figure}[ht]
     \centering
     \begin{subfigure}[b]{0.195\textwidth}
         \centering
         \includegraphics[width=\textwidth]{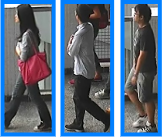}
         \caption{CUHK-03}
         \label{fig:imageCUHK}
     \end{subfigure}
     \hfill
     \begin{subfigure}[b]{0.26\textwidth}
         \centering
         \includegraphics[width=\textwidth]{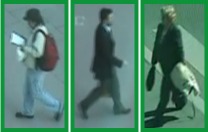}
         \caption{PRID 2011}
         \label{fig:imagePRID}
     \end{subfigure}
     
     \caption{Example images from the CUHK-03 and the PRID-2011 datasets.}
     \label{fig:imagesComparison}
\end{figure}

The fact that a network trained on CUHK-03 can generalize to data from another dataset shows that the proposed Re-ID model is able to learn cross-domain Re-ID. Indeed, the kind of images used for training are very different than the images encountered at inference time (see Fig.~\ref{fig:imagesComparison}). This property is interesting as the domains encountered for every new implementation vary a lot with cameras' resolution, distance the people and illumination, among other factors.

\begin{figure*}
     \centering
     \begin{subfigure}[b]{0.32\textwidth}
         \centering
         \includegraphics[width=\textwidth]{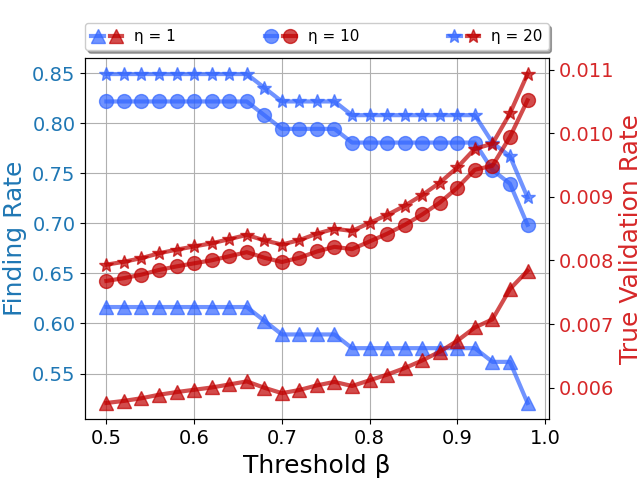}
         \caption{$\tau$=10 }
         \label{fig:fr_with_t10}
     \end{subfigure}
     \hfill
     \begin{subfigure}[b]{0.32\textwidth}
         \centering
         \includegraphics[width=\textwidth]{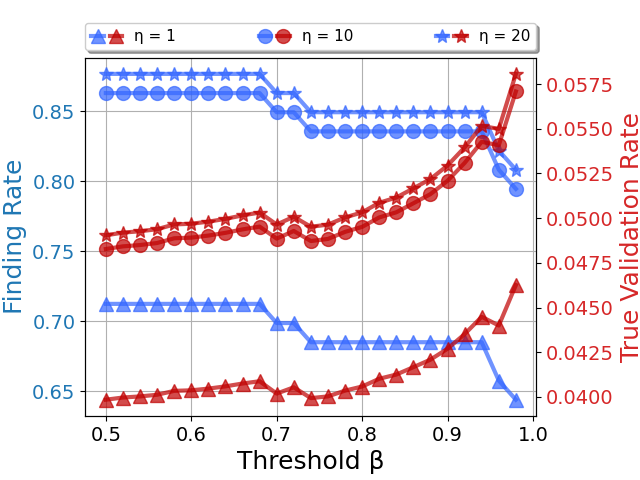}
         \caption{$\tau$=100}
         \label{fig:fr_with_t100}
     \end{subfigure}
     \hfill
     \begin{subfigure}[b]{0.32\textwidth}
         \centering
         \includegraphics[width=\textwidth]{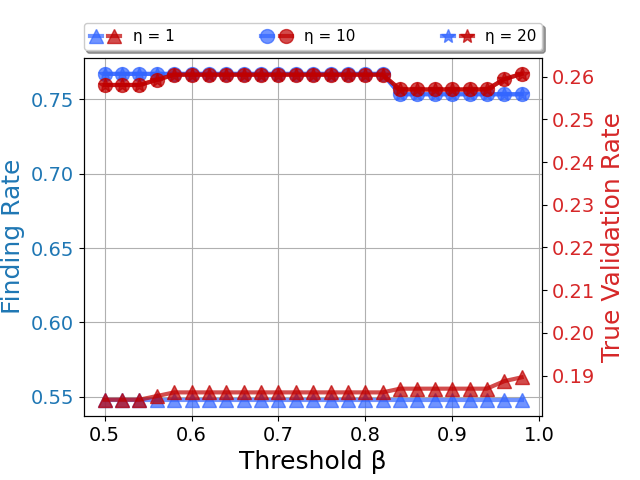}
         \caption{$\tau$=1000}
         \label{fig:fr_with_t1000}
     \end{subfigure}
 \caption{Finding Rate (FR) and True Validation Rate (TVR) curves for different values of $\tau$, $\beta$ and $\eta$. \textit{In Fig.~\ref{fig:fr_with_t1000}, the curves $\eta=10$ and $\eta=20$ are overlapping}.}
 \label{fig:results graphs}
\end{figure*}

\subsection{Evaluation of the full pipeline for FF-PRID}

For evaluation, 10 videos of two minutes were selected from each view. For each video, approximately 4 query images were selected. In total, our evaluation consists of 20 videos and 73 queries. Each query appears in its associated video at least in one frame, but does not necessarily appear in each sub-videos after splitting into shorter sequences. To evaluate the influence of the different parameters of the FF-PRID pipeline, i.e. the number of frames for video splitting $\tau$, the threshold for alert generation $\beta$ and the number of candidates shown to the monitoring agent $\eta$, we use different values for each parameter. Thus, we test $\tau \in \{10, 100, 1000\}$, $\eta \in \{1, 10, 20\}$ and the threshold $\beta$ is computed for various values in the interval $\left[0.5, 0.98\right]$. Fig.~\ref{fig:results graphs} shows the FR and TVR curves for different values of $\tau$, $\beta$ and $\eta$.

\subsubsection{Influence of the FF-PRID parameters}

In Fig.~\ref{fig:results graphs}, for all values of $\tau$ and $\eta$, the FR curves decrease when $\beta$ increases. This behavior is due to the fact that with larger $\beta$, the system raises less alerts and is more likely to miss the query. However, with $\tau = 1000$, the decreasing effect is less noticeable. This is because when considering larger galleries, the model has more chances of finding a similar image and having at least one high confidence prediction. 
In contrast, the three TVR curves are increasing with $\beta$. This also makes sense as increasing $\beta$ correspond to reducing the accepted confidence range and thus calling the agent less frequently. However, except for $\tau = 1000$, we note that the values of the different TVR are all very low, meaning that the human monitoring agent would be called in many unnecessary cases. 

Furthermore, as expected, $\eta = 10$ and $\eta = 20$ performed much better than $\eta = 1$ for all configurations of $\tau$ and $\beta$. Indeed, the C-PRID models are not perfect and training Re-ID models with very high top 1 accuracy is hard. In contrast, decreasing $\eta$, reduces the amount of work for the monitoring agent as it needs to control less image samples. 
Besides, the FR curves present better results for $\tau = 100$ than for the two other tested values, meaning that the raw video is split into sub-videos which are neither too short nor too long. This way, the query appears on the video for a sufficient amount of time to be recognize and there are not too many distractors to confuse the network.  

\subsubsection{Qualitative Evaluation}

\begin{figure}[ht]
     \centering
     \includegraphics[width=0.4\textwidth]{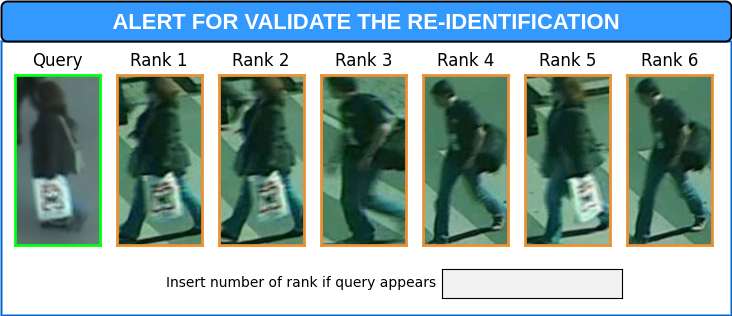}
     \caption{Example of the interface for alert validation with $\eta=6$.}
     \label{fig:interfaceReID}
\end{figure}

Fig.~\ref{fig:interfaceReID} shows an example of the propositions received by the monitoring agent. 
In practice, during the evaluation, we observed that way too many alert calls. Also, when a sub-video of $\tau$ frames contains too many persons, a variety of cropped images are presented, not always representing the query. We also noted that the imperfect BBs produced by YOLO-v3 (Fig.~\ref{fig:mistakesCropps}) had a strong negative influence on the results.

\subsubsection{Further considerations}

The results obtained for the FF-PRID problem suggest that careful selection of the tunable parameters ($\tau$, $\beta$ and $\eta$) is paramount. Indeed, with proper selection we can reach an FR of almost 80\% with a TVR of 26\%. Although the score that we managed to reach for the Finding Rate are satisfactory, we acknowledge that the TVR is still too low for the method to be used practically, as the operator would be called too many times if dealing with several cameras at the same time. These mixed results emphasize the importance of considering the FF-PRID problem as a whole and suggest that changing the paradigm for person Re-ID might be the best way to obtain applicable solution for tomorrow's cities.

As already mentioned, the discrepancy between the training domain of the classic Re-ID model and the domain generated by OD is a possible reason for the results obtained. Another possible reason for the low TVR is that SiamIDL is a closed-world method, i.e. it supposes that the query is always present in the search gallery. Hence, the highest ranked prediction tend to have very high confidence scores and to raise alerts very frequently. In the validation example studied here, sometimes the query is not present, thus defining an open set scenario~\cite{openworld}.

\section{Conclusion and Future Work}
\label{sec:conclusion}


In this paper we claim that the classic approach for person Re-ID is not sufficient to develop robust practical implementations for security application. Indeed, full frames of the cameras stream must be processed instead of pre-cropped clean images of people. 
To study FF-PRID, we present a new framework, taking into account the fact that a human monitoring agent needs to validate the results. We also introduce two new metrics in order to evaluate FF-PRID models within the proposed pipeline. The metrics are used to evaluate how many times the query is found when it is present in the video (Finding Rate) and how many times the query is present when the agent is solicited (True Validation Rate). These framework and metrics are, to the best of our knowledge, the first proposed approaches to evaluate a FF-PRID model, looking for persons directly in the entire video frames.

To support our claim, we conducted experiments on a modified version of the PRID-2011 dataset, using a standard pipeline for FF-PRID. First, persons BBs are extracted from the input video using a state-of-the-art OD model to generate a search gallery. Then, the query is searched in the gallery using a good C-PRID model. We demonstrated that both the OD model and the C-PRID models managed to perform well on the new dataset without additional training. However, the final results for FF-PRID, evaluated using FR and TVR, were not sufficient to deploy FF-PRID in production. Although choosing the right parameters in our framework enabled us to reach a good FR score ($>80\%$), we were not able to obtain a TVR much better than $25\%$, which means that most of the time the operator calls were unnecessary. Some possible explanations for these results were discussed as well as possible improvements. These mixed results emphasize the importance of considering Re-ID in the FF-PRID setting if we want to develop methods that can be used reliably in practical scenarios. We believe that many improvements could be achieved if the community starts investigating Re-ID solutions for the Full Frame setting instead of focusing only on the classic pre-cropped image-based setting.

After demonstrating the importance of considering the FF-PRID setting, the next step is to propose new approaches to improve the standard pipeline implemented here, and get closer to solving FF-PRID. A possible direction to achieve this is to consider video-based classic Re-ID methods~\cite{PRL_videoBased, PRL_videoBased2}. Another natural option is to consider the open-world person Re-ID setting instead of closed-world~\cite{openworld}. We also plan to train more specific pedestrian detection techniques, focusing on recognizing only full-bodies. Finally, another potential improvement to address the problem would consist in building a large dataset of annotated videos, which could be used for training an end-to-end model for the whole FF-PRID application. This approach sounds promising in regards with the success of end-to-end approaches in solving complex tasks lately~\cite{end2end_1, end2end_2}.



\end{document}